\documentclass[10pt,twocolumn,letterpaper]{article}

\usepackage{cvpr}
\usepackage{times}
\usepackage{epsfig}
\usepackage{graphicx}
\usepackage{amsmath}
\usepackage{amssymb}
\usepackage{xcolor}         
\usepackage{float}
\usepackage{multirow}
\usepackage{graphics}
\usepackage{graphicx}
\usepackage{amsmath}
\usepackage{diagbox}
\usepackage{cancel}

\newcommand{\datasetname}{OASIS}
\newcommand{\datasetsize}{140,000}

\usepackage[breaklinks=true,bookmarks=false]{hyperref}

\cvprfinalcopy 

\def\cvprPaperID{} 

\ifcvprfinal\fi
\begin{document}

\title{\datasetname: A Large-Scale Dataset for Single Image 3D in the Wild}

\author{	
Weifeng Chen\textsuperscript{1,2}\kern10pt Shengyi Qian\textsuperscript{1}\kern10pt David Fan\textsuperscript{2}\kern10pt Noriyuki Kojima\textsuperscript{1}\kern10pt Max Hamilton\textsuperscript{1}\kern10pt Jia Deng\textsuperscript{2} 
\vspace{2mm}
\\
\begin{minipage}{\columnwidth}
	\centering
	\textsuperscript{1}University of Michigan, Ann Arbor
	{\tt\small \{wfchen,syqian,kojimano,johnmaxh\}@umich.edu}\\
\end{minipage}
\begin{minipage}{\columnwidth}
	\centering
	\textsuperscript{2}Princeton University\\
	{\tt\small dfan@alumni.princeton.edu,jiadeng@princeton.edu}\\
\end{minipage}  
}

\makeatletter
\g@addto@macro\@maketitle{
\begin{figure}[H]
\setlength{\linewidth}{\textwidth}
\setlength{\hsize}{\textwidth}
\centering
\includegraphics[width=\linewidth]{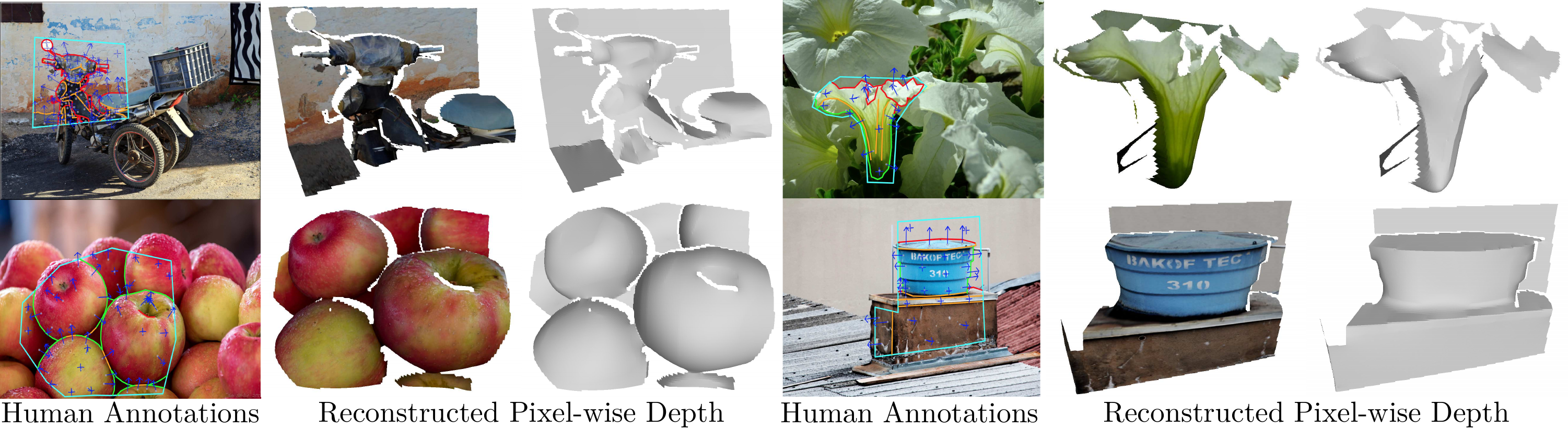}
\vspace{-2mm}
\caption{We introduce Open Annotations of Single-Image Surfaces ({\datasetname}), a large-scale dataset of human annotations of 3D surfaces for {\datasetsize} images in the wild. More examples in the appendix.}
\label{fig:teaser}
\end{figure}
}
\makeatother

\maketitle

\begin{abstract}
Single-view 3D is the task of recovering 3D properties such as depth and surface normals from a single image. We hypothesize that a major obstacle to single-image 3D is data. We address this issue by presenting Open Annotations of Single Image Surfaces ({\datasetname}), a dataset for single-image 3D in the wild consisting of annotations of detailed 3D geometry for {\datasetsize} images. We train and evaluate leading models on a variety of single-image 3D tasks. We expect {\datasetname} to be a useful resource for 3D vision research. Project site: \url{https://pvl.cs.princeton.edu/OASIS}. 
\end{abstract}

\vspace{-5mm}

\section{Introduction}

Single-view 3D is the task of recovering 3D properties such as depth and surface normals from a single RGB image. It is a core computer vision problem of critical importance. 3D scene interpretation is a foundation for understanding events and planning actions. 3D shape representation is crucial for making object recognition robust against changes in viewpoint, pose, and illumination. 3D from a single image is especially important due to the ubiquity of monocular images and videos. Even with a stereo camera with which 3D can be reconstructed by triangulating matching pixels from different views, monocular 3D cues are still necessary in textureless or specular regions where it is difficult to reliably match pixel values. 

Single-image 3D is challenging. Unlike multiview 3D, it is ill-posed and resists tractable analytical formulation except in the most simplistic settings. As a result, data-driven approaches have shown greater promise, as evidenced by a plethora of works that train deep networks to map an RGB image to depth, surface normals, or 3D models~\cite{eigen2014depth,laina2016deeper,wang2016surge,ilg2018occlusions,zeisl2014discriminatively,mescheder2018occupancy}. However, despite substantial progress, the best systems today still struggle with handling scenes ``in the wild''--- arbitrary scenes that a camera may encounter in the real world. As prior work has shown~\cite{chen2016single}, state-of-art systems often give erroneous results when presented with unfamiliar scenes with novel shapes or layouts. 

We hypothesize that a major obstacle of single-image 3D is data. Unlike object recognition, whose progress has been propelled by datasets like ImageNet~\cite{deng2009imagenet} covering diverse object categories with high-quality labels, single-image 3D has lacked an ImageNet equivalent that covers diverse scenes with high-quality 3D ground truth. Existing datasets are restricted to either a narrow range of scenes~\cite{silberman2012indoor,dai2017scannet} or simplistic annotations such as sparse relative depth pairs or surface normals~\cite{chen2016single,chen2017surface}. 

In this paper we introduce \emph{Open Annotations of Single-Image Surfaces} ({\datasetname}), a large-scale dataset for single-image 3D in the wild. It consists of human annotations that enable pixel-wise reconstruction of 3D surfaces for {\datasetsize} randomly sampled Internet images. Fig.~\ref{fig:teaser} shows the human annotations of example images along with the reconstructed surfaces. 

A key feature of {\datasetname} is its rich annotations of human 3D perception. Six types of 3D properties are annotated for each image: occlusion boundary (depth discontinuity), fold boundary (normal discontinuity), surface normal, relative depth, relative normal (orthogonal, parallel, or neither), and planarity (planar or not). These annotations together enable a reconstruction of pixelwise depth. 

To construct {\datasetname}, we created a UI for interactive 3D annotation. The UI allows a crowd worker to annotate the aforementioned 3D properties. It also provides a live, rotatable rendering of the resulting 3D surface reconstruction to help the crowd worker fine-tune their annotations.

It is worth noting that {\datasetsize} images may not seem very large compared to millions of images in datasets like ImageNet. But the number of images  can be a misleading metric. For {\datasetname}, annotating one image takes 305 seconds on average. In contrast, verifying a single image-level label takes no more than a few seconds. Thus in terms of the total amount of human time, {\datasetname} is already comparable to millions of image-level labels. 

{\datasetname} opens up new research opportunities on a wide range of single-image 3D tasks---depth estimation, surface normal estimation, boundary detection, and instance segmentation of planes---by providing in-the-wild ground truths either for the first time, or at a much larger scale than prior work. For depth estimation and surface normals, \emph{pixelwise} ground truth is available for images in the wild for the first time---prior data in the wild provide only sparse annotations~\cite{chen2016single,chen2019learning}. For the detection of occlusion boundaries and folds, {\datasetname} provides annotations at a scale 700 times larger than prior work---existing datasets~\cite{stein2009occlusion,Karsch:CVPR:13} have annotations for only about 200 images. For instance segmentation of planes, ground truth annotation is available for images in the wild for the first time. 

To facilitate future research, we provide extensive statistics of the annotations in  {\datasetname}, and train and evaluate leading deep learning models on a variety of single-image tasks. Experiments show that there is a large room for performance improvement, pointing to ample research opportunities for designing new learning algorithms for single-image 3D. We expect {\datasetname} to serve as a useful resource for 3D vision research. 

\section{Related Work}

\noindent\textbf{3D Ground Truth from Depth-Sensors and Computer Graphics}
Major 3D datatsets are either collected by sensors~\cite{silberman2012indoor,Geiger2013IJRR,
saxena20083,schops2017multi,dai2017scannet} or synthesized with Computer Graphics
~\cite{Butler:ECCV:2012,mccormac2016scenenet,song2016ssc,MIFDB16,richter2016playing}. But due
to the limitations of depth sensors and the lack of varied 3D assets for 
rendering, the diversity of scenes is quite limited. For example,  
sensor-based ground truth is mostly for indoor or 
driving scenes~\cite{silberman2012indoor,dai2017scannet,mccormac2016scenenet,song2016ssc,Geiger2013IJRR}. 

\smallskip
\noindent\textbf{3D Ground Truth from Multiview Reconstruction} Single-image 3D training data can also be obtained by applying classical Structure-from-Motion (SfM) algorithms on Internet images or videos~\cite{li2018megadepth,xian2018monocular,chen2019learning}. 
However, classical SfM algorithms have many well known failure modes including scenes with moving objects and scenes with specular or textureless surfaces. In contrast, humans can annotate all types of scenes. 

\smallskip
\noindent\textbf{3D Ground Truth from Human Annotations}
Our work is connected to many previous works
that crowdsource 3D annotations of Internet images. For example, prior work has crowdsourced annotations of relative depth~\cite{chen2016single} and surface normals~\cite{chen2017surface} at sparse locations of an image (a single pair of relative depth and a single normal per image). 
Prior work has also aligned pre-existing 3D models to images~\cite{xiang2016objectnet3d,sun2018pix3d}. However, this approach has a drawback that not every shape can be perfectly aligned with available 3D models, whereas our approach can handle arbitrary geometry.

Our work is related to that of Karsch et al.~\cite{Karsch:CVPR:13}, who
reconstruct pixelwise depth from human annotations of boundaries, with the aid of a shape-from-shading algorithm~\cite{barron2012color}. 
Our approach is different in that  
we annotate not only boundaries but also surface normals, planarity, and relative normals, and our reconstruction method does not rely on automatic shape from shading, which is still unsolved and has many failure modes.

One of our inspirations is LabelMe3D~\cite{russell2009building}, which annotated 3D planes attached to a common ground plane.  
Another is OpenSurfaces~\cite{bell13opensurfaces}, which also annotated 3D planes. 
We differ from LabelMe3D and OpenSurfaces in that our annotations recover not only
planes but also curved surfaces. 
Our dataset is also much larger, being $600\times$ the size of LabelMe3D 
and $5\times$ of OpenSurfaces in terms of the number of images annotated. It is also more diverse, because LabelMe3D and OpenSurface 
include only city or indoor scenes.

\begin{figure*}[t]
\begin{center}
\includegraphics[width=\linewidth]{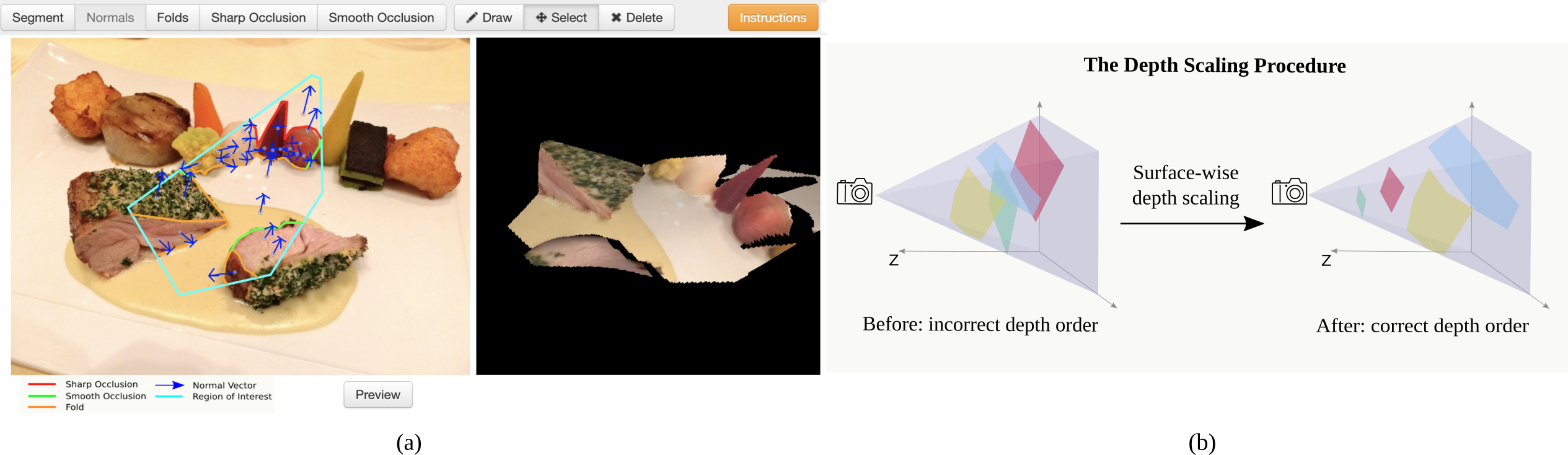}
\end{center}
\vspace{-3mm}
\caption{\textbf{(a)} Our UI allows a user to annotate rich 3D properties and includes a preview window for interactive 3D visualization. \textbf{(b)} An illustration of the depth scaling procedure in our backend. }
\label{fig:pipeline}
\end{figure*}

\section{Crowdsourcing Human Annotations}
\label{sec:data_annotation}

We use random keywords to query and download Creative Commons Flickr images with a known focal length (extracted from the EXIF data). Each image is presented to a crowd worker for annotation through a custom UI  as shown in Fig.~\ref{fig:pipeline} (a).  The worker is asked to mask out a region 
that she wishes to work on with a polygon of her choice, with the requirement that the polygon covers a pair of randomly pre-selected locations. She then works on the annotations and iteratively monitors
the generated mesh (detailed in Sec~\ref{sec:dense_depth}) from an interactive preview window (Fig.~\ref{fig:pipeline} (a)). 

\smallskip
\noindent\textbf{Occlusion Boundary and Fold} An occlusion boundary denotes locations of depth discontinuity, where the surface on 
one side is physically disconnected from the surface on the other side. When it is drawn, the worker
also specifies which side of the occlusion is closer to the viewer, i.e. depth order of the surfaces on both sides of the occlusion. Workers need to distinguish between two kinds 
of occlusion boundaries. \emph{Smooth occlusion} (green in Fig~\ref{fig:pipeline} (a)) is where the the closer surface smoothly curves away from the viewer, and the surface 
normals should be orthogonal to the occlusion line and parallel to the image plane, and 
pointing toward the further side. \emph{Sharp occlusion} (red in Fig~\ref{fig:pipeline} (a)) has none of these constraints.
On the other hand, \emph{fold} denotes locations of surface normal discontinuity, where the surface geometry changes abruptly, but the surfaces
on the two sides of the fold are still physically attached to each other (orange in Fig~\ref{fig:pipeline} (a)). 

Occlusion boundaries segment a region into subregions, each of which is a \emph{continuous surface} whose geometry can change abruptly but remains physically connected in 3D. Folds further segment a continuous surface into \emph{smooth surfaces} where the geometry vary smoothly without discontinuity of surface normals.  

\smallskip
\noindent\textbf{Surface Normal} The worker first specifies if a smooth surface is planar or curved. 
She annotates one normal at each planar surface which indicates the orientation of the plane.  
For each curved surface, she annotates normals at as many locations as she sees fit.
A normal is visualized as a blue arrow originating from a green grid (see the appendix), 
rendered in perspective projection according to the known focal length. 
Such visualization helps workers perceive the normal in 3D~\cite{chen2017surface}. To rotate and 
adjust the normal, the worker only needs to drag the mouse.

\smallskip
\noindent\textbf{Relative Normal} Finally, to annotate normals with higher accuracy, 
the worker specifies the \emph{relative normal} between each 
pair of planar surfaces. She chooses between \emph{Neither}, \emph{Parallel} and \emph{Orthogonal}. 
Surfaces pairs that are parallel or orthogonal to each other then have their normals adjusted 
automatically to reflect the relation.

\smallskip
\noindent\textbf{Interactive Previewing} While annotating, the worker can click a button to see a visualization of the 3D shape constructed from the current
annotations (detailed later in Sec.~\ref{sec:dense_depth}). Workers can rotate or zoom to inspect the shape from different angles in
a preview window (Fig~\ref{fig:pipeline} (a)). 
She keeps working on it until she is satisfied with the shape.

\smallskip
\noindent\textbf{Quality Control}
Completing our 3D annotation task requires knowledge of relevant concepts.
To ensure good quality of the dataset, we require each worker to complete a 
training course to learn concepts such as occlusions, folds and normals, and usage of the UI. 
She then needs to pass a qualification quiz before being allowed to work on our annotation task. Besides explicitly selecting qualified workers, we also set up a separate quality verification task on each collected mesh. In this task, a worker inspects the mesh to judge if it
reflects the image well. Only meshes deemed high quality are accepted.

To improve our annotation throughput, we collected annotations from three sources: Amazon Mechanical Turk, which accounts for 11\% of all annotations, and two data annotation companies that employ full-time annotators, who supplied the rest of the annotations.

\begin{figure*}[ht!]
\begin{center}
	\includegraphics[width=\linewidth]{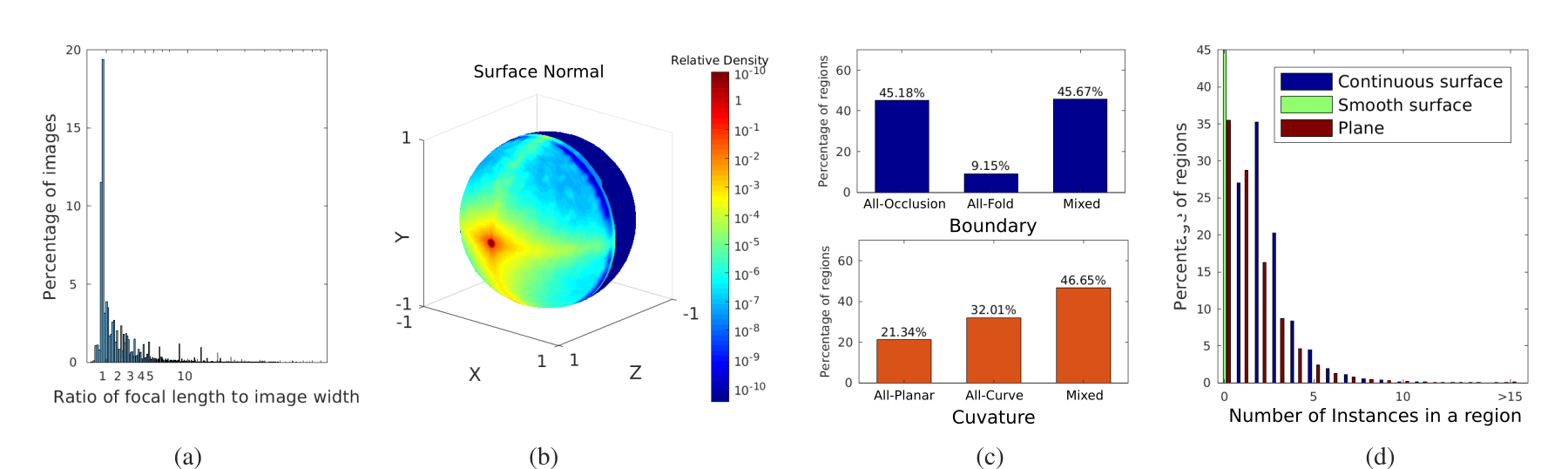}
\end{center}
\vspace{-5mm}
\caption{Statistics of {\datasetname}. (a) The distribution of focal length (unit: relative length to the image width). (b) The distribution of surface normals. (c) Boundary: the ratio of regions containing only occlusion, only fold, and both. Curvature: the distribution of regions containing only planes, only curved surfaces, and both. (d) The frequency distribution of each surface type in a region.}
\label{fig:all_fig}
\end{figure*}

\section{From Human Annotations to Dense Depth}
\label{sec:dense_depth}

Because humans do not directly annotate the depth value of each pixel, we need to convert the human annotations to pixelwise depth in order to visualize the 3D surface. 

\smallskip
\noindent\textbf{Generating Dense Surface Normals}
We first describe how we generate dense surface normals from annotations. 
We assume the normals to be smoothly varying in the spatial domain, except across folds or 
occlusion boundaries where the normals change abruptly. 
Therefore, our system propagates the known normals to the unknown ones by requiring the 
final normals to be smooth overall, but stops the propagation at fold and occlusion lines.

More concretely, let $N_p$ denote the normal at pixel $p$ on a normal map $N$, 
and $F$, $O$  denotes the pixels belong to the folds and occlusion boundaries. 
We have a set of known normals $\tilde{N}$ at locations $P_{known}$ from 
(1) surface normal annotations by workers, and 
(2) the pre-computed normals along the smooth occlusion boundaries as mentioned 
in Sec~\ref{sec:data_annotation}. 
Each pixel $p$ has four neighbors $\Phi(p)$. 
If $p$ is on an occlusion boundary, its neighbors on the 
closer side of this boundary are $\Gamma_O(p)$. 
If $p$ is on a fold line, only its neighbors $\Gamma_F(p)$
on one fixed random side of this line are considered. We solve for
the optimal normal $N^*$ using LU factorization
and then normalize it into unit norm:

\vspace{-5mm}

\begin{equation}
\begin{aligned}
N^* & = \operatorname*{argmin}_N  
\sum_{p \not\in F\cup O}\sum_{\substack{q\in \Phi(p) \\ q \not\in F\cup O} }|N_p-N_q|^{2} +  \\
&\sum_{p \in O}\sum_{\substack{q\in \Gamma_O(p)} }|N_p-N_q|^{2} +
\sum_{p \in F}\sum_{\substack{q\in \Gamma_F(p)} }|N_p-N_q|^{2}\\
\end{aligned}
\end{equation}
\begin{align}
\text{s.t.} & \quad N_p = \tilde{N}_p, \forall p \in P_{known}
\end{align}

\smallskip
\noindent\textbf{Generating Dense Depth} Our depth generation pipeline consists of two stages: First, from surface normals and focal length,
we recover the depth of each \emph{continuous surface} through integration~\cite{queau2018normal}
\footnote{We snap the z component of the surface normals to be no smaller than 0.3 so that the generated depth would not stretch into huge distance.}.
Next, we adjust the depth order among these surfaces by performing surface-wise depth scaling
(Fig.~\ref{fig:pipeline} (b)), i.e. each surface has its own scale factor. 

Our design is motivated by this fact: in single-view depth recovery,
depth within continuous surface can be recovered only up to an ambiguous scale;
thus different surfaces may end up with different scales, leading to incorrect 
depth ordering between surfaces. 
But workers already decide which side of an occlusion boundary is closer to the viewer. 
Based on such knowledge, we correct depth order
by scaling the depth of each surface.

We now describe the details. Let $\mathbf{S}$ denotes the set of all continuous surface.
From integration, we obtain the depth $Z_S$ of each $S\in\mathbf{S}$. 
We then solve for a scaling factor $X_S$ for each $S$, which is used in scaling depth $Z_{S}$. 
Let $\mathbf{O}$ denote the set of occlusion boundaries.
Along $\mathbf{O}$, we densely sample a set of point pairs $\mathbf{B}$. 
Each pair $(p,q) \in \mathbf{B}$ has $p$ lying on the closer side of one of the occlusion boundaries
$O_i \in \mathbf{O}$ and $q$ the further side. 
The continuous surface a pixel $p$ lies on is $S(p)$, and its depth is $Z_p$.
The set of optimal scaling factors $\mathbf{X^*}$ is solved for as follows:
\begin{align}
\mathbf{X^*} & =  \operatorname*{argmin}_{\mathbf{X}}   
\sum_{S\in\mathbf{S}} X_{S} \label{eq:min_sum}\\
\text{s.t.} & \quad X_{S(p)} Z_{p} + \epsilon \leq X_{S(q)} Z_{q}, \forall (p,q) \in \mathbf{B} \label{eq:constraint}\\
& \quad X_{S} \geq \eta, \forall S \in \mathbf{S}
\end{align}
where $\epsilon > 0$ is a minimum separation between surfaces, and $\eta > 0$ is a minimum scale factor. Eq.(\ref{eq:constraint}) requires the surfaces to meet the depth order constraints
specified by point pairs $(p,q) \in \mathbf{B}$ after scaling. Meanwhile, Eq.(\ref{eq:min_sum})
constrains the value of $\mathbf{X}$ so that they do not increase indefinitely. 
After correcting the depth order, the final depth for surface $S$ is $X_{S}^* Z_{S}$.
We normalize and reproject the final depth to 3D as point clouds, and generate 3D meshes for visualization. 

\begin{table*}[ht!]
\centering
\resizebox{\textwidth}{!}{\begin{tabular}{ |c | c c c | c c c|} 
        \hline      
                       & \multicolumn{3}{c|}{NYU Depth~\cite{silberman2012indoor} (depth mean: 2.471 m, depth std: 0.754 m)} &    \multicolumn{3}{c|}{Tanks \& Temples~\cite{Knapitsch2017} (depth mean: 4.309m, depth std: 3.059m)}                  \\
        \cline{2-7}    
        	                 &    Human-Human     &    Human-Sensor        &          CNN-Sensor                       &       Human-Human    &    Human-Sensor       &    CNN-Sensor       \\
        \hline				
Depth (EDist)               &    0.078m          &      0.095m            &          0.097m~\cite{laina2016deeper}    &       0.194m         &     0.213m                   &      0.402m~\cite{laina2016deeper}         \\
     Normals (MAE)           &    13.13$^{\circ}$ &     17.82$^{\circ}$    &14.19$^{\circ}$~\cite{zhang2016physically} &     14.33$^{\circ}$  &    20.29$^{\circ}$    &  29.11$^{\circ}$~\cite{zhang2016physically}\\
Post-Rotation Depth (EDist)    &    0.037m          &      0.048m            &            -                              &       0.082m         &     0.080m            &         -                                  \\
    Depth Order (WKDR)       &     5.68\%         &      8.67\%            &          11.90\%                          &       9.28\%         &     10.80\%           &      32.13\%                               \\
        \hline
	\end{tabular}}
\vspace{1mm}
\caption{Depth and normal difference between different humans (Human-Human), between human and depth sensor (Human-Sensor), and between ConvNet and depth sensor (CNN-Sensor). The results are averaged over all human pairs.}
\label{tab:annotation_accuracy}
\end{table*}

\begin{figure*}[ht!]
\begin{center}
\includegraphics[width=\linewidth]{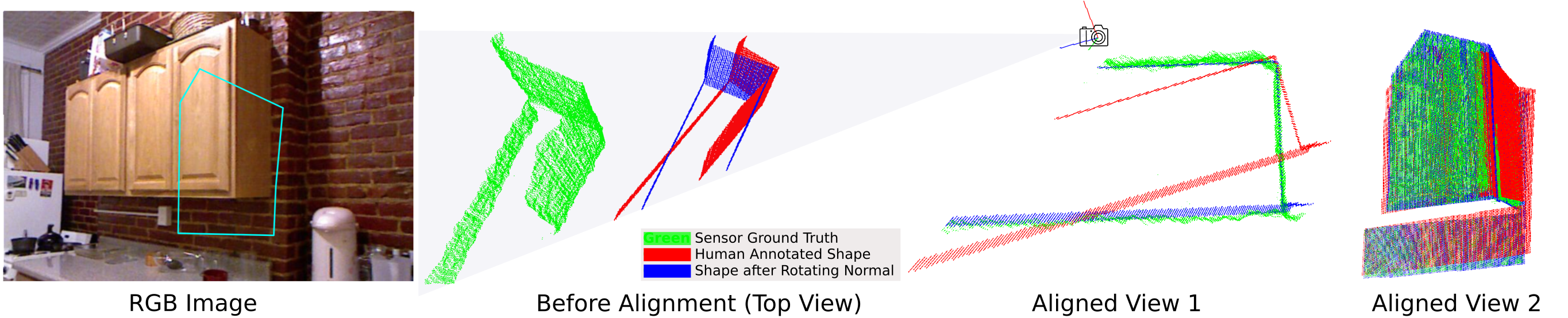}
\end{center}
\vspace{-5mm}
\caption{Humans estimate shape correctly but the absolute orientation can be slightly off, causing large depth error after perspective back-projection into 3D. Depth error drops significantly (from 0.07m to 0.01m) after a global rotation of normals.} 
\label{fig:orientation}
\end{figure*}

\section{Dataset Statistics}

\smallskip
\noindent\textbf{Statistics of Surfaces}
Fig.~\ref{fig:all_fig} plots various statistics of the 3D surfaces. 
Fig.~\ref{fig:all_fig} (a) plots the distribution of focal length. We see that focal lengths in {\datasetname} vary greatly: they range from wide angle to telezoom, and are mostly 1$\times$ to 10$\times$ of the width of the image. 
Fig.~\ref{fig:all_fig} (b) visualizes the distribution of surface normals. We see that a substantial proportion of normals point directly towards the camera, suggesting that parallel-frontal surfaces frequently occur in natural scenes. 
Fig.~\ref{fig:all_fig} (c) presents region-wise statistics. 
We see that most regions (90\%+) contain occlusion boundaries and close to half have both occlusion boundaries and folds (top). 
We also see that most regions (70\%+) contain at least one curve surface (bottom).
Fig.~\ref{fig:all_fig} (d) shows the histogram of the number of different kinds of surfaces in an annotated region. 
We see that most regions consist of multiple disconnected pieces and have non-trivial geometry in terms of continuity and smoothness.

\smallskip
\noindent\textbf{Annotation Quality}
We study how accurate and consistent the annotations are. To this end, we randomly sample 50 images from NYU Depth~\cite{silberman2012indoor}
and 70 images from Tanks and Temples~\cite{Knapitsch2017}, and have 20 workers annotate each image. 
Tab.~\ref{tab:annotation_accuracy} reports the depth and normal difference between human annotations, 
between human annotations and sensor ground truth, and between predictions from state-of-the-art ConvNets and sensor ground truth. 
Depth difference is measured by the mean Euclidean distance (EDist) between corresponding points in two point clouds, 
after aligning one to the other through a global translation and scaling (surface-wise scaling for human annotations and CNN predictions). Normal difference is measured in Mean Angular Error (MAE). 
We see in Tab.~\ref{tab:annotation_accuracy} that human annotations are highly consistent with each other and with sensor ground truth, 
and are better than ConvNet predictions, especially when the ConvNet is not trained and tested on the same dataset.  

We observe that humans often estimate the shape correctly, but the overall orientation can be slightly off, causing a large depth error
against sensor ground truth (Fig.~\ref{fig:orientation}). This error can be particularly pronounced for planes close to orthogonal to the image plane. Thus we also compute the error after a rotational alignment with the sensor ground truth---we globally rotate the human annotated normals (up to 30 degrees) before generating the shape. After accounting for this global rotation of normals, human-sensor depth difference is further reduced by 47.96\% (relative) for NYU and 62.44\% (relative)
for Tanks and Temples; a significant drop of normal error is also observed in human-human difference.

We also measure the qualitative aspect of human annotations by evaluating the WKDR metric~\cite{chen2016single}, 
i.e. the percentage of point pairs with inconsistent depth ordering between query and reference depth.
Depth pairs are sampled in the same way as~\cite{chen2016single}. 
Tab.~\ref{tab:annotation_accuracy} again shows that human annotations are qualitatively accurate and highly consistent with each other.

It is worth noting that metric 3D accuracy is not required for many tasks such as navigation, object manipulation, and semantic scene 
understanding---humans do well without perfect metric accuracy. Therefore human perception of depth alone can be the gold standard for training
and evaluating vision systems, regardless of its metric accuracy. As a result, our dataset would still be valuable even if it were less 
metrically accurate than it is currently.

\begin{figure*}[t]
\begin{center}
	\includegraphics[width=\linewidth]{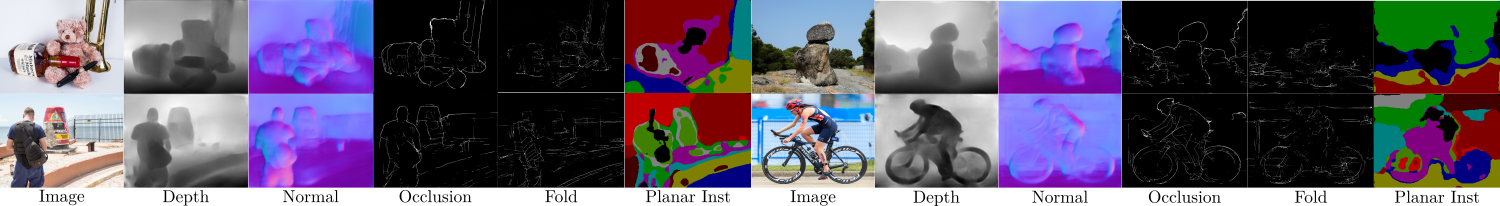}
\end{center}
\vspace{-5mm}
\caption{Qualitative outputs of the four tasks from representative models.
	More details and examples are in the appendix.}
\label{fig:qual_examples}
\end{figure*}

\section{Experiments}

To facilitate future research, we use {\datasetname} to  train and evaluate leading deep learning models on a suite of single-image 3D tasks including depth estimation, normal estimation, boundary detection, plane segmentation. Qualitative results are shown in Fig.~\ref{fig:qual_examples}. A train-val-test split of 110K, 10K, 20K is used for all tasks. 

For each task we estimate human performance to provide an upperbound accounting for the variance of human annotations. We randomly sample 100 images from the test set, and have each image re-annotated by 8 crowd workers. That is, each image now has ``predictions'' from 8 different humans. We evaluate each prediction and report the mean as the performance expected of an average human.

\vspace{-1mm}

\subsection{Depth Estimation}
\vspace{-1mm}
We first study single-view depth estimation.
{\datasetname} provides pixelwise \emph{metric} depth in the wild.
But as discussed in Sec~\ref{sec:dense_depth}, due to inherent single-image ambiguity, depth in {\datasetname} is independently recovered within each continuous 
surface, after which the depth undergoes a surface-wise scaling to correct the depth order. 
The recovered depth is only accurate up to scaling within each continuous surface and ordering between continuous surfaces.

\def\mycmd{2}

Given this, in {\datasetname} we provide metric depth ground truths that is surface-wise accurate up to a scaling factor.
This new form of depth necessitates new evaluation metrics and training losses.

\smallskip
\noindent\textbf{Depth Metric} 
The images in {\datasetname} have varied focal lengths. 
This means that to evaluate depth estimation, we cannot simply use pixelwise difference between a predicted depth map and the ground truth map. This is because the predicted 3D shape depends greatly on the focal length---given the same depth values, decreasing the focal length will flatten the shape along the depth dimension. In practice, the focal length is often unknown for a test image. Thus, we require a depth estimator to predict a focal length along with depth. Because the predicted focal length may differ from the ground truth focal length, pixelwise depth difference is a poor indicator of how close the predicted 3D shape is to the ground truth. 

A more reasonable metric is the
Euclidean distance between the predicted and ground-truth 3D point cloud. 
Concretely, we backproject the predicted depth $Z$ to a 3D point cloud $\mathbf{P} = \{(X_p,Y_p,Z_p)\}$ using $f$ (the predicted focal length), 
and ground truth depth $Z^*$ to $\mathbf{P^*} = \{(X^*_p,Y^*_p,Z^*_p)\}$ using $f^*$ (the ground truth focal length). 
We then calculate the distance between $\mathbf{P}$ and $\mathbf{P^*}$.

The metric also needs to be invariant to surface-wise depth scaling and translation.
Therefore we introduce a surface-wise scaling factor $\lambda_{S_i} \in \mathbf{\Lambda}$, and a surface-wise translation $\delta_{S_i} \in \mathbf{\Delta}$,  
to align each predicted surface $S_i \in \mathbf{S}$ in $\mathbf{P}$ to the ground truth point cloud  $\mathbf{P^*}$ in a least square manner. 
The final metric, which we call Locally Scale-Invariant RMSE (LSIV\_RMSE), is defined as:

\vspace{-5mm}

\begin{equation}
\begin{aligned}
LSIV\_RMS&E (Z, Z^*) = \operatorname*{min}_{\mathbf{\Lambda},\mathbf{\Delta}}   
\sum_{p} (  \frac{(X^*_p, Y^*_p, Z^*_p)}{\sigma(X^*)}  \\ 
& - \lambda_{S(p)}(X_p, Y_p, Z_p) - (0,0,\delta_{S(p)})  )^2,  \\
\end{aligned}
\end{equation}
where $S(p)$ denotes the surface a pixel $p$ is on.
The ground truth point cloud $\mathbf{P^*}$ is normalized to a canonical scale by the standard deviation 
of its X coordinates $\sigma(X^*)$.
Under this metric, as long as $\mathbf{P}$ is accurate up to scaling and translation, it will align 
perfectly with $\mathbf{P^*}$, and get 0 error.

Note that LSIV\_RMSE ignore the ordering between two separate surfaces; it allows objects floating in the air to be arbitrarily scaled. This is typically not an issue because in most scenes there are not many objects floating in the air. But we nonetheless also measure the correctness of depth ordering. We report WKDR~\cite{chen2016single}, 
which is the percentage of point pairs that have incorrect depth order in the predicted depth.  
We evaluate on depth pairs sampled in the same way as~\cite{chen2016single}, i.e. half are random pairs, 
half are from the same random horizontal lines.

\smallskip
\noindent\textbf{Models}
We train and evaluate two leading depth estimation networks on {\datasetname}: 
the Hourglass network~\cite{chen2016single}, and ResNetD~\cite{xian2018monocular}, a dense prediction network based on ResNet50. 
Each network predicts a metric depth map and a focal length, which are together used to backproject pixels to 3D points, which are compared against the ground truth to compute the LSIV\_RMSE metric, which we optimize as the loss function during training. Note that we do not supervise on the predicted focal length. 

We also evaluate leading pre-trained models that estimate single-image depth on {\datasetname}, including  
FCRN~\cite{laina2016deeper} trained on ILSVRC~\cite{russakovsky2015imagenet} and NYU Depth~\cite{silberman2012indoor}, 
Hourglass~\cite{li2018megadepth} trained on MegaDepth~\cite{li2018megadepth}, ResNetD~\cite{xian2018monocular} trained on 
a combination of datasets including ILSVRC~\cite{russakovsky2015imagenet}, Depth in the Wild~\cite{chen2016single}, ReDWeb~\cite{xian2018monocular} and YouTube3D~\cite{chen2019learning}. 
For networks that do not produce a focal length, we use the validation set to find the best focal length 
that leads to the smallest LSIV\_RMSE, and use this focal length for each test image. In addition, we also evaluate \emph{plane}, a naive baseline that predicts a uniform depth map.

\begin{table}[h]
\begin{center}
	
	\resizebox{\columnwidth}{!}{ \begin{tabular}{|c |c |c c |}
			\hline      
		            Method                     &                  Training Data                                               &     LSIV\_RMSE        &           WKDR                      \\
			\hline	
FCRN~\cite{laina2016deeper}                    &     ImageNet~\cite{russakovsky2015imagenet} + NYU~\cite{silberman2012indoor} &  0.67 (\cancel{0.67}) &       39.95\%   (\cancel{39.94\%})  \\
Hourglass~\cite{chen2016single, li2018megadepth}&     MegaDepth~\cite{li2018megadepth}                                        &  0.67 (\cancel{0.67}) &       38.37\%   (\cancel{38.37\%})\\
\multirow{2}{*}{ResNetD~\cite{xian2018monocular,chen2019learning}}&     ImageNet~\cite{russakovsky2015imagenet} + YouTube3D~\cite{chen2019learning}+&\multirow{2}{*}{0.66 (\cancel{0.66})} &\multirow{2}{*}{34.01\% (\cancel{34.03\%})} \\
                                               &     ReDWeb~\cite{xian2018monocular} +  DIW~\cite{chen2016single}             &                       &              \\
\hline
ResNetD~\cite{xian2018monocular}               &     ImageNet~\cite{russakovsky2015imagenet} + {\datasetname}                 &  0.37 (\cancel{0.37}) &       32.62\%  (\cancel{32.04\%})   \\
ResNetD~\cite{xian2018monocular}               &     {\datasetname}                                                           &  0.47 (\cancel{0.47}) &       39.73\%  (\cancel{38.79\%})   \\
Hourglass~\cite{chen2016single}                &     {\datasetname}                                                           &  0.45 (\cancel{0.47}) &       39.01\%  (\cancel{39.64\%})   \\
\hline
                        Plane                  &              -                                                               &  0.67 (\cancel{0.67}) &       100.00\%  (\cancel{100.00\%}) \\
                     Human (Approx)            &              -                                                               &  0.24 (\cancel{0.24}) &        19.04\%  (\cancel{19.33\%})  \\
			\hline
	\end{tabular}}
\end{center}
\caption{Depth estimation performance of different networks on {\datasetname} (lower is better). For networks that do not produce a focal length, we use the best focal length leading to the smallest error. See Sec. A\ref{appendix:crossed_out_numbers} about the numbers crossed out. }
\label{table:depth_performance}
\end{table}

Tab.~\ref{table:depth_performance} reports the results. In terms of metric depth, we see that networks trained on OASIS perform the best. This is expected because they are trained to predict a focal length and to directly optimize the LSIV\_RMSE metric. It is noteworthy that ImageNet pretraining provides a significant benefit even for this purely geometrical task. Off-the-shelf models do not perform better than the naive baseline, probably because they were not trained on diverse enough scenes or were not trained to optimize metric depth error. In terms of relative depth, it is interesting to see that ResNetD trained on ImageNet and OASIS performs the best, even though the training loss does not enforce depth ordering. We also see that there is still a significant gap between human performance and machine performance. At the same time, the gap is not hopelessly large, indicating the effectiveness of a large training set.

\begin{table}[ht!]
\begin{center}
	\resizebox{\columnwidth}{!}{
		\begin{tabular}{| c | c | c  c | c  c  c | c c |}
			\hline      
			                                      &                               & \multicolumn{7}{c|}{\datasetname} \\ \cline{3-9}
			Method	                              & Training Data                 &  \multicolumn{2}{c|}{Angle Distance}& \multicolumn{3}{c|}{\% Within $t^{\circ}$}  &   \multicolumn{2}{c|}{Relative Normal}     \\
			                                      &                               &Mean                    &  Median                & 11.25$^{\circ}$        & 22.5$^{\circ}$         & 30$^{\circ}$           & $AUC_{o}$               & $AUC_{p}$\\
            \hline				
            Hourglass~\cite{chen2017surface}      &  {\datasetname}               & 23.91 (\cancel{23.24}) & 18.16 (\cancel{18.08}) & 31.23 (\cancel{31.44}) & 59.45 (\cancel{59.79}) & 71.77 (\cancel{72.25}) & 0.5913(\cancel{0.5508}) &  0.5786 (\cancel{0.5439}) \\				
            Hourglass~\cite{chen2017surface}      & SNOW~\cite{chen2017surface}   & 31.35 (\cancel{30.74}) & 26.97 (\cancel{26.65}) & 13.98 (\cancel{14.33}) & 40.20 (\cancel{40.84}) & 56.03 (\cancel{56.73}) & 0.5329(\cancel{0.5329}) &  0.5016 (\cancel{0.4714}) \\
            Hourglass~\cite{chen2017surface}      &NYU~\cite{silberman2012indoor} & 35.32 (\cancel{34.69}) & 29.21 (\cancel{28.76}) & 14.23 (\cancel{14.65}) & 37.72 (\cancel{38.49}) & 51.31 (\cancel{52.06}) & 0.5467(\cancel{0.5415}) &  0.5132 (\cancel{0.5064}) \\				
            PBRS~\cite{zhang2016physically}       &NYU~\cite{silberman2012indoor} & 38.29 (\cancel{38.09}) & 33.16 (\cancel{33.00}) & 11.59  (\cancel{11.94})& 32.14 (\cancel{32.58}) & 45.00 (\cancel{45.29}) & 0.5669(\cancel{0.5729}) &  0.5253 (\cancel{0.5227})\\
            \hline
            Front\_Facing                         &             -                 & 31.79 (\cancel{31.20}) & 24.80 (\cancel{24.76}) & 27.52 (\cancel{27.36}) & 46.61 (\cancel{46.62}) & 56.80 (\cancel{56.94}) & 0.5000 (\cancel{0.5000}) & 0.5000 (\cancel{0.5000})\\
            Human (Approx)                        &             -                 & 17.27 (\cancel{17.43}) & 12.92 (\cancel{13.08}) & 44.36 (\cancel{43.89}) & 76.16 (\cancel{75.94}) & 85.24 (\cancel{84.72}) & 0.8826 (\cancel{0.8870}) & 0.6514 (\cancel{0.6439})\\
		    \hline
		    
		\end{tabular}
	}
\end{center}
\caption{Surface normal estimation on {\datasetname}. See Sec. A\ref{appendix:crossed_out_numbers} about the numbers crossed out.}
\label{table:OASIS_performance}
\end{table}

\begin{table}[ht!]
\begin{center}
	\resizebox{\columnwidth}{!}{
		\begin{tabular}{| c | c | c  | c  c  c |c | c  c  c   |}
			\hline      
			                &                                    & \multicolumn{4}{c|}{DIODE~\cite{vasiljevic2019diode}} &  \multicolumn{4}{c|}{ETH3D~\cite{schops2017multi}}  \\ \cline{3-10}
			Method	        &       Training Data                & Angle Distance& \multicolumn{3}{c|}{\% Within $t^{\circ}$} & Angle Distance& \multicolumn{3}{c|}{\% Within $t^{\circ}$}   \\
			                &                                    & Mean & 11.25$^{\circ}$ & 22.5$^{\circ}$ & 30$^{\circ}$& Mean & 11.25$^{\circ}$ & 22.5$^{\circ}$ & 30$^{\circ}$ \\
			\hline				
Hourglass~\cite{chen2017surface}& {\datasetname}                 & 34.21 (\cancel{34.57}) & 14.45 (\cancel{13.71}) & 36.98 (\cancel{35.69}) & 51.36 (\cancel{49.65}) & 33.00 (\cancel{34.51}) &  26.25(\cancel{23.52}) & 54.07 (\cancel{52.04}) & 65.36 (\cancel{62.73}) \\				
Hourglass~\cite{chen2017surface}& SNOW~\cite{chen2017surface}    & 40.10 &  8.29 & 27.20 & 40.67 & 45.71 & 10.69 & 31.16 & 43.16 \\
Hourglass~\cite{chen2017surface}& NYU~\cite{silberman2012indoor} & 42.23 & 10.97 & 29.76 & 41.35 & 41.84 & 21.94 & 44.05 & 53.81 \\
PBRS~\cite{zhang2016physically} & NYU~\cite{silberman2012indoor} & 42.59 &  9.96 & 29.08 & 40.72 & 39.91 & 18.68 & 44.76 & 56.08 \\
			Front\_Facing   &               -                    & 47.76 &  5.62 & 18.70 & 28.05 & 58.97 & 11.84 & 23.75 & 30.19 \\
			\hline
		\end{tabular}
	}
\end{center}
\caption{Cross-dataset generalization. See Sec. A\ref{appendix:crossed_out_numbers} about the numbers crossed out.}
\label{table:cross_dataset_normal_performance}
\end{table}

\subsection{Surface Normal Estimation}

We now turn to single-view surface normal estimation. 
We evaluate on absolute normal, i.e. the pixel-wise predicted normal values, and
\emph{relative normal}, i.e. the parallel and orthogonal relation predicted between planar surfaces.

\smallskip
\noindent\textbf{Absolute Normal Evaluation} 
We use standard metrics proposed in prior work~\cite{wang2015designing}: the mean and median of angular error measured 
in degrees, and the percentage of pixels whose angular error is within $\gamma$ degrees.

We evaluate on {\datasetname} four state-of-the-art networks that are trained to directly predict normals:  
(1) Hourglass~\cite{chen2017surface} trained on {\datasetname}, 
(2) Hourglass trained on the Surface Normal in the Wild (SNOW) dataset~\cite{chen2017surface}, 
(3) Hourglass trained on NYU Depth~\cite{silberman2012indoor}, and
(4) PBRS, a normal estimation network by Zhang et al.~\cite{zhang2016physically} trained on NYU Depth~\cite{silberman2012indoor}. We also include 
Front\_Facing, a naive baseline predicting all normals to be orthogonal to the image plane. 

\begin{figure}[t]
\begin{center}
	\includegraphics[width=\columnwidth]{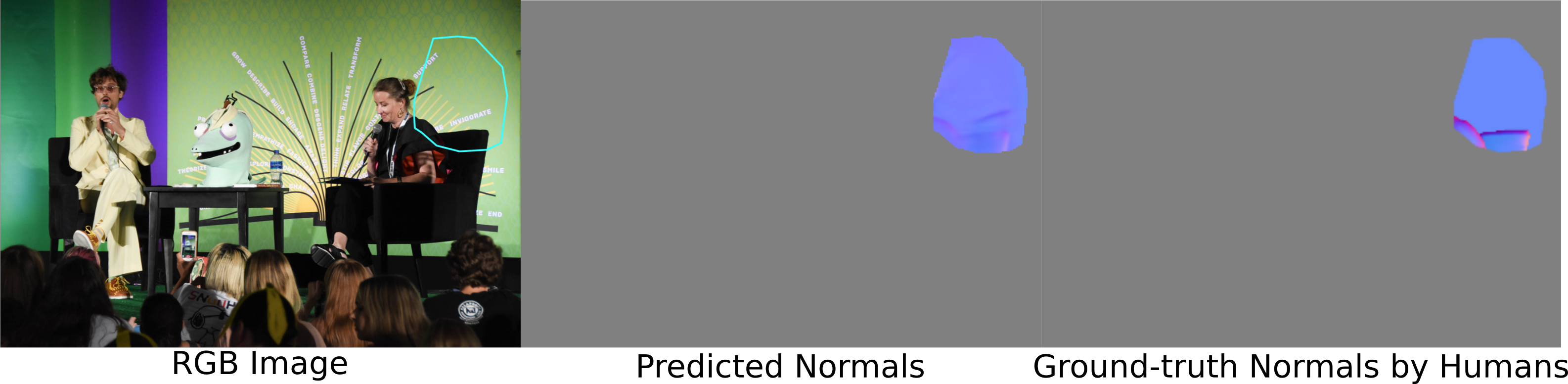}
\end{center}
\vspace{-4mm}
\caption{Limitations of standard metrics: a deep network gets low mean angle error but important details are wrong.}
\label{fig:normal_metric_error}
\end{figure}

Tab.~\ref{table:OASIS_performance} reports the results. As expected, the Hourglass network trained on {\datasetname} performs the best. Although SNOW is also an in-the-wild dataset, 
the same network trained on it does not perform as well, but is still better than training on NYU. Notably, the human-machine gap appears fairly small numerically (17.27 versus 23.91 in mean angle error). However, we observe that the naive baseline can achieve 31.79; thus the dynamic range of this metric is small to start with, due to the natural distribution of normals in the wild. In addition, a close examination of the results suggests that these standard metrics of surface normals do not align well with perceptual quality. In natural images there can be large areas that dominate the metric but have uninteresting geometry, such as a blank wall in the background. For example, in Fig.~\ref{fig:normal_metric_error}, a neural network gets the background correct, but largely misses the important details in the foreground. This opens up an interesting research question about developing new evaluation metrics.

\smallskip
\noindent\textbf{Relative Normal Evaluation} 
We also evaluate the predicted normals in terms of relative relations, specifically orthogonality and parallelism. Getting these relations correct is important because it can help find vanishing lines and perform self-calibration.

We first define a metric to evaluate relative normal. 
From the human annotations, we first sample an equal number of point pairs from
surface pairs that are parallel, orthogonal, and neither. 
Given a predicted normal map, we look at the two normals at each point pair and 
measure the angle $\theta$ between them. 
We consider them orthogonal if $|cos(\theta - 90^\circ)| < cos(\Theta_{o})$, and parallel
if $|cos(\theta)| > cos(\Theta_{p})$, where $\Theta_{o}$, $\Theta_{p}$ are thresholds. 
We then plot the Precision-and-Recall curve for orthogonal by varying $\Theta_{o}$,
and measure its Area Under Curve $\mathbf{AUC_{o}}$, using \emph{neither} and \emph{parallel} pairs as negative examples. Varying $\Theta_{p}$ and using \emph{neither} and \emph{orthogonal} as negative examples, we obtain $\mathbf{AUC_{p}}$ for parallel.

Tab.~\ref{table:OASIS_performance} reports results of relative normal evaluation. 
Notably, all methods perform similarly, and all perform very poorly compared to humans. This suggests that existing approaches to normal estimation have limitations in capturing orthogonality and parallelism, indicating the need for further research.

\smallskip
\noindent\textbf{Cross-Dataset Generalization} 
Next we study how networks trained on {\datasetname} generalize to other datasets. Surface normal estimation is ideal for such evaluation because unlike depth, which is tricky to evaluate on a new dataset due to scale ambiguity and varying focal length, a normal estimation network can be directly evaluated on a new dataset without modification. 

We train the same Hourglass network on {\datasetname}, and NYU, and report their performance on two benchmarks not seen in training: DIODE~\cite{vasiljevic2019diode} and ETH3D~\cite{schops2017multi}. From Tab.~\ref{table:cross_dataset_normal_performance} 
we see that training on NYU underperforms on all benchmarks, showing that networks trained on 
scene-specific datasets have difficulties generalizing to diverse scenes. 
Training on {\datasetname} outperforms on all benchmarks, demonstrating the effectiveness of diverse annotations.

\subsection{Fold and Occlusion Boundary Detection}

Occlusion and fold are both important 3D cues, as they tell us about physical connectivity and curvature: 
\emph{Occlusion} delineates the boundary at which surfaces are physically disconnected to each other, 
while \emph{Fold} is where geometry changes abruptly but the surfaces remain connected. 

\smallskip
\noindent\textbf{Task} We investigate joint boundary detection and occlusion-versus-fold classification: deciding whether a pixel is a boundary (fold or occlusion) and if so, which kind it is.
Prior work has explored similar topics: Hoiem et al.~\cite{hoiem2011recovering}
and Stein et al.~\cite{stein2009occlusion} handcraft edge or motion features to perform occlusion 
detection, but our task involves folds, not just occlusion lines.

\begin{table}[h!]
\begin{center}
	\resizebox{\columnwidth}{!}{ \begin{tabular}{| c | c   c  c  c  |c|  }
			\hline
\diagbox{Metric}{Model}	    & Edge: All Fold &  Edge: All Occ  & HED~\cite{xie2015holistically}      &  Hourglass~\cite{chen2016single}  & Human (Approx) \\
			\hline				
			ODS             &   0.123        &     0.539       &      0.547      (\cancel{0.533})    &     0.581        (\cancel{0.585}) &   0.810        \\
			OIS             &   0.129        &     0.576       &      0.606      (\cancel{0.584})    &     0.639        (\cancel{0.639}) &   0.815        \\
			AP              &    0.02        &      0.44       &      0.488      (\cancel{0.466})    &     0.530        (\cancel{0.547}) &   0.642        \\
			\hline
	\end{tabular}}
\end{center}
\caption{Boundary detection performance on {\datasetname}. See Sec. A\ref{appendix:crossed_out_numbers} about the numbers crossed out.}
\label{table:occ_fold}
\end{table}

\smallskip
\noindent\textbf{Evaluation Metric} We adopt metrics similar to standard ones 
used in edge detection~\cite{arbelaez2011contour,xie2015holistically}: F-score by optimal 
threshold per image (OIS), by fixed threshold (ODS) and average precision (AP). 
For a boundary to be considered correct, it has to be labeled correctly as either occlusion or fold. More details on the metrics can be found in the 
appendix.

To perform joint detection of fold and occlusion, we adapt and train two networks on {\datasetname}: 
Hourglass~\cite{chen2016single}, and 
a state-of-the-art edge detection network HED~\cite{xie2015holistically}.
The networks take in an image, and output two probabilities per pixel: $p_e$ is the probability of 
being an boundary pixel (occlusion or fold), and $p_f$ is the probability of being a fold pixel. Given a 
threshold $\tau$, pixels whose $p_e < \tau$ are neither fold nor occlusion.
Pixels whose $p_e > \tau$ are fold if $p_F > 0.5$ and otherwise occlusion. 

As baselines, we also investigate how a generic edge detector would perform on this task. 
We use HED network trained on BSDS dataset~\cite{arbelaez2011contour} to detect image edges, 
and classify the resulting edges to be either all occlusion (\textit{Edge: All Occ}) or all 
fold (\textit{Edge: All Fold}). 

All results are reported on Tab~\ref{table:occ_fold}.
Hourglass outperforms HED when trained on {\datasetname}, and significantly outperforms both the All-Fold and All-Occlusion baselines, but still underperforms humans by a large margin, 
suggesting that fold and occlusion boundary detection remains challenging in the wild.

\vspace{-1mm}

\subsection{Instance Segmentation of Planes}

Our last task focuses on instance segmentation of planes in the wild.
This task is important because planes often have special functional roles in a scene (e.g.\@ supporting surfaces, walls). 
Prior work has explored instance segmentation of planes, but is limited to indoor or driving environments~\cite{liu2018planenet,yu2019single,liu2018planercnn,yang2018recovering}.
Thanks to {\datasetname}, we are able to present the first-ever evaluation of this task in the wild. 

We follow the way prior work~\cite{liu2018planenet,liu2018planercnn,yu2019single} performs this task:  
a network takes in an image, and produces instance masks of planes, 
along with an estimate of planar parameters that define each 3D plane. 
To measure performance, we report metrics used in instance segmentation literature~\cite{lin2014microsoft}:
the average precision (AP) computed and averaged across a range of overlap thresholds (ranges from 50\% to 95\% as in~\cite{lin2014microsoft,cordts2016cityscapes}). 
A ground truth plane is considered correctly detected if 
it overlaps with one of the detected planes by more than the overlap threshold, 
and we penalize multiple detection as in~\cite{cordts2016cityscapes}.
We also report the AP at 50\% overlap ($\text{AP}^{50\%}$) and 75\% overlap ($\text{AP}^{75\%}$).

PlanarReconstruction by Yu et al.~\cite{yu2019single} is a state-of-the-art method for planar instance segmentation. We train PlanarReconstruction on three combinations of data:  (1) ScanNet~\cite{dai2017scannet} only as done in~\cite{yu2019single}, (2) {\datasetname} only, 
and (3) ScanNet + {\datasetname}. 
Tab.~\ref{table:instance_\datasetname} compares their performance.

As expected, training on ScanNet alone performs the worse, because ScanNet only has indoor images.
Training on {\datasetname} leads to better performance. Leveraging both ScanNet and {\datasetname} is the best overall. 
But even the best network significantly underperforms humans, suggesting ample space for improvement.

\begin{table}[h!]
\begin{center}
	\resizebox{\columnwidth}{!}{\begin{tabular}{| c | c | c | c | c |   }
			\hline      
			Method                   & Training Data                                             &       AP                  &  $\text{AP}^{50\%}$            &  $\text{AP}^{75\%}$ \\
			\hline	
			                         & ScanNet~\cite{dai2017scannet}                             &  0.076 (\cancel{0.076})    &     0.161 (\cancel{0.161})     & 0.064 (\cancel{0.065})       \\
PlanarReconstruction~\cite{yu2019single} & {\datasetname}                                        &  0.125 (\cancel{0.127})    &     0.249 (\cancel{0.250})     & 0.110 (\cancel{0.112})       \\
			                         & ScanNet~\cite{dai2017scannet} + {\datasetname}            &  0.137 (\cancel{0.139})    &     0.262 (\cancel{0.264})     & 0.126 (\cancel{0.130})       \\
			                         \hline
Human (Approx)                       &                           -                               &      0.461                &     0.542                      &      0.476          \\
			\hline
\end{tabular}}
\end{center}
\caption{Planar instance segmentation performance on {\datasetname}. See Sec. A\ref{appendix:crossed_out_numbers} about the numbers crossed out.
}
\label{table:instance_\datasetname}
\end{table}

\vspace{-3mm}

\section{Conclusion}
We have presented {\datasetname}, a dataset
of rich human 3D annotations. We trained and evaluated leading models on a variety of single-image tasks. We expect {\datasetname} to be a useful resource for 3D vision research.

\smallskip
\noindent\textbf{Acknowledgement}
This work was partially supported by a National Science Foundation grant (No. 1617767), a Google gift, and a Princeton SEAS innovation grant.

{\small
\bibliographystyle{ieee_fullname}
\bibliography{egbib}
}


\begin{center}
\textbf{{\LARGE Appendix}}
\end{center}

\renewcommand{\thefigure}{\Roman{figure}}
\renewcommand{\thetable}{\Roman{table}}
\setcounter{figure}{0} 
\setcounter{table}{0} 

\section*{A1: Surface Normal Annotation UI}

\begin{figure}[h!]
	\begin{center}
		\includegraphics[width=\columnwidth]{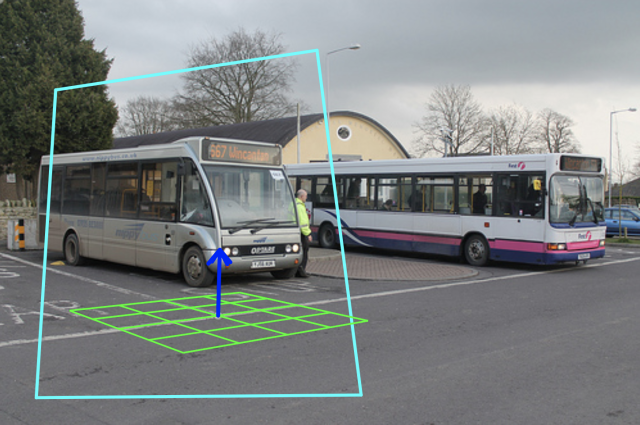}
	\end{center}
	\caption{Surface normal annotation UI. The surface normal is visualized as a blue arrow originating from a green grid, 
rendered in perspective projection according to the known focal length.}
	\label{fig:normal_ui}
\end{figure}

The surface normal annotation UI is shown in Fig.~\ref{fig:normal_ui}. 

\section*{A2: Planar versus Curved Regions}

Tab.~\ref{tab:annotation_accuracy} measures the annotation quality separately for planar regions and curved regions. 

\begin{table}[ht!]
\centering
\resizebox{0.7\columnwidth}{!}{\begin{tabular}{ | c | c c |} 
        \hline      
                             & \multicolumn{2}{c|}{NYU Depth~\cite{silberman2012indoor}}  \\
        \cline{2-3}    
        	                 &    Human-Human     &    Human-Sensor        \\
        \hline				
Planar Regions                &     0.079m         &      0.091m             \\
Curved Regions                &     0.077m         &      0.102m             \\
        \hline
	\end{tabular}}
\vspace{1mm}
\caption{Depth difference between different humans (Human-Human) and between humans and depth sensors (Human-Sensor) in planar and curved regions. The results are averaged over all human pairs.  The mean of depth in tested samples is 2.471 m, the standard deviation is 0.754 m.}
\label{tab:annotation_accuracy}
\end{table}

\section*{A3: Comparison with Other Datasets}

Tab.~\ref{tab:dataset_comparison} compares {\datasetname} and other datasets. 

\begin{table*}[ht!]
    \centering
    \resizebox{\textwidth}{!}{
        \begin{tabular}{c|ccccccccc}
            Dataset                     & In the Wild     & Acquisition     & Depth              & Normals & Occlusion  \& Fold        & Relative Normals           & Planar Inst Seg &  \# Images \\ 
            \hline
   {\datasetname}                       & \checkmark      & Human annotation & Metric (up to scale) & Dense   & \checkmark                & \checkmark                & \checkmark      &  140K      \\
NYU Depth V2~\cite{silberman2012indoor} & -               & Kinect          & Metric       & Dense   & -                         & -                         & -               &   407K      \\
KITTI~\cite{Geiger2013IJRR}             & -               & LiDAR           & Metric       & -       & -                         & -                         & -               &   93K       \\
DIW~\cite{chen2016single}               & \checkmark      & Human annotation & Relative     & -       & -                         & -                         & -               &   496K      \\
SNOW~\cite{chen2017surface}             & \checkmark      & Human annotation & -                  & Sparse  & -                         & -                         & -               &   60K       \\
MegaDepth~\cite{li2018megadepth}        & \checkmark      & SfM             & Metric (up to scale) & -       & -                         & -                         & -               &   130K      \\
ReDWeb~\cite{xian2018monocular}         & \checkmark      & Stereo          & Metric (up to scale) & -       & -                         & -                         & -               &   3.6K      \\
3D Movie~\cite{lasinger2019towards}     & \checkmark      & Stereo          & Metric (up to scale) & -       & -                         & -                         & -               &   75K       \\
OpenSurfaces~\cite{bell13opensurfaces}  & -               & Human annotation & -                  & Dense   & -                         & -                         & -               &   25K       \\
CMU Occlusion~\cite{stein2009occlusion}  & \checkmark     & Human annotation & -                  & -       & Occlusion Only            & -                         & -               &   538       \\
        \end{tabular}}
	\vspace{1mm}
    \caption{Comparison between OASIS and other 3D datasets. \emph{Metric (up to scale)} denotes that the depth is metrically accurate up to scale.}
    \label{tab:dataset_comparison}
\end{table*}

\begin{figure*}[ht!]
	\begin{center}
		\includegraphics[width=0.95\linewidth]{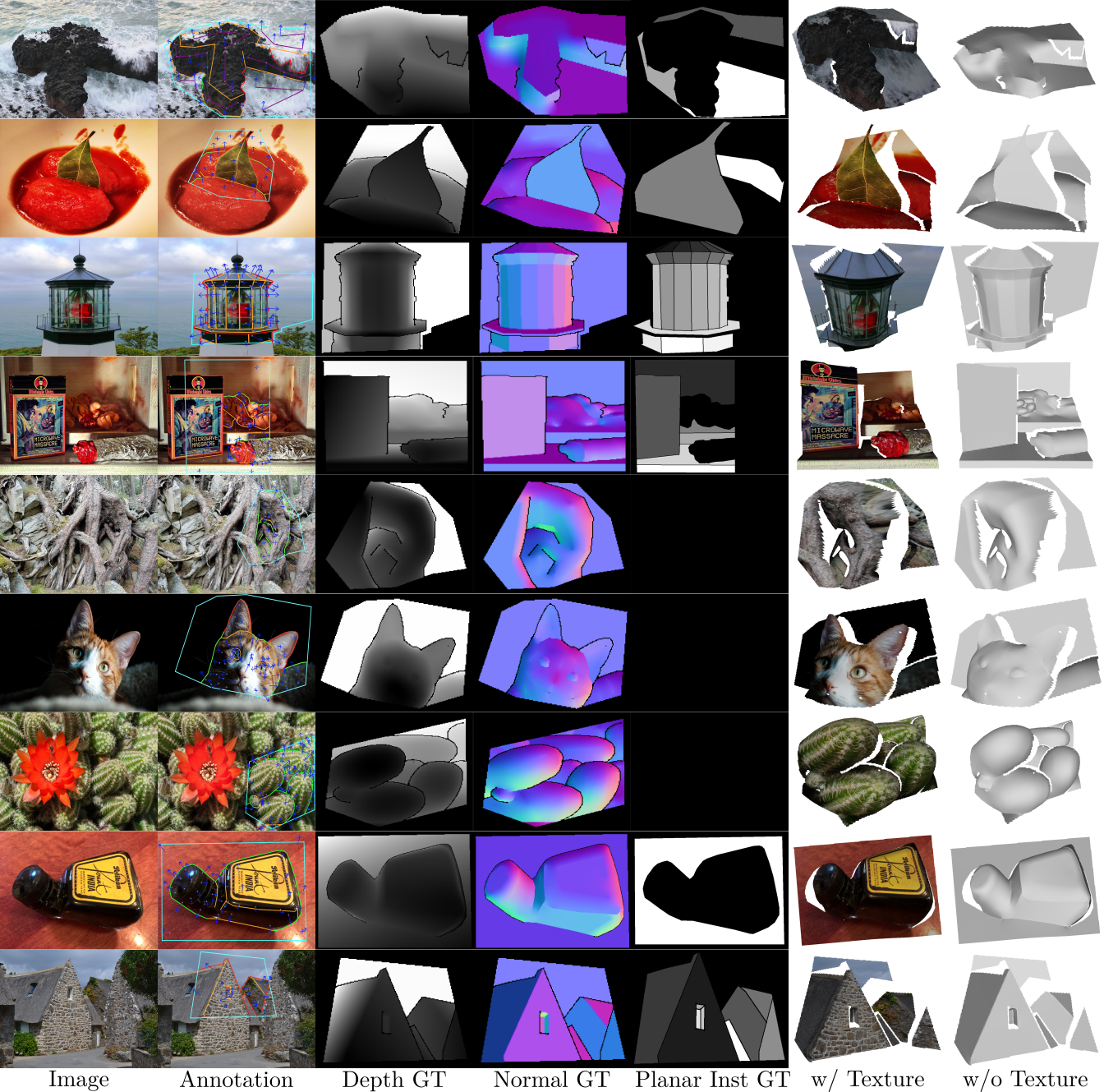}
	\end{center}
	\vspace{-3mm}
	\caption{Additional human annotations from {\datasetname}. Note that each planar instance has a different color. }
	\label{fig:supp_teaser}
\end{figure*}

\vspace{-3mm}
\section*{A4: Additional Examples from {\datasetname}}

Additional human annotations are shown in Fig.~\ref{fig:supp_teaser}.

\section*{A5: Additional Qualitative Outputs}
\begin{figure*}[h]
	\begin{center}
		\includegraphics[width=0.83\linewidth]{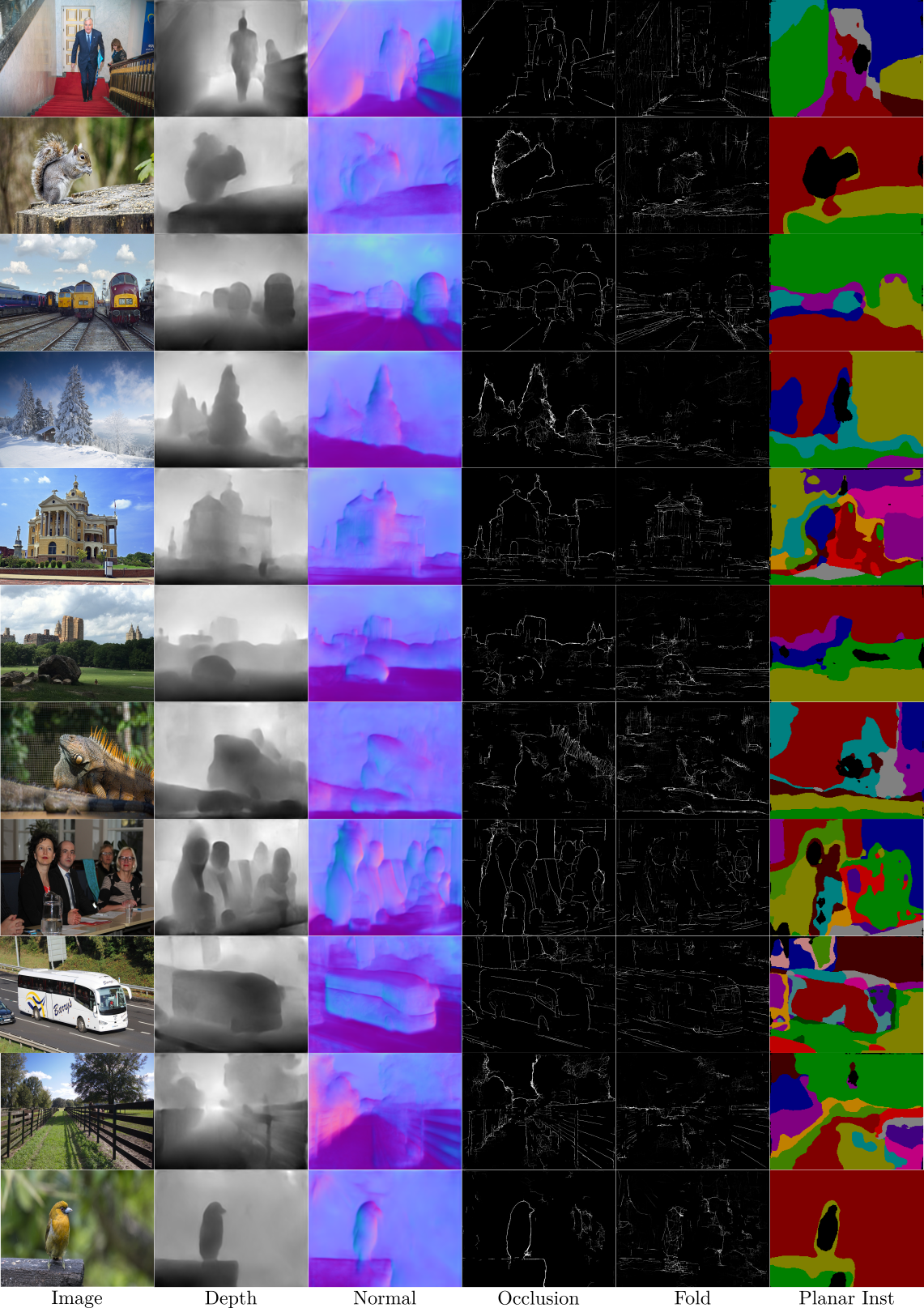}
	\end{center}
	\caption{Additional qualitative outputs from four tasks: (1) depth estimation, (2) normal estimation, (3) fold and occlusion boundary detection, and (4) planar instance segmentation. }
	\label{fig:supp_qual}
\end{figure*}

Qualitative predictions presented in both Fig.~\ref{fig:supp_qual} 
and Fig.~\ref{fig:qual_examples} are produced as follows:
Depth predictions are produced by a ResNetD~\cite{xian2018monocular} network trained on {\datasetname} + ImageNet~\cite{deng2009imagenet}.
Surface normal predictions are produced by an Hourglass~\cite{chen2017surface} network trained on {\datasetname} alone.
Occlusion boundary and fold predictions are produced by an Hourglass~\cite{chen2016single} network trained on {\datasetname} alone.
Planar instance segmentations are produced by a PlanarReconstruction~\cite{yu2019single} network trained on Scannet~\cite{dai2017scannet} + {\datasetname}.

\section*{A6: Evaluating Fold and Occlusion Boundary Detection}
This section provides details on evaluating fold and occlusion boundary detection. As discussed in Sec 6.3 of the main paper, our metric is based on the ones used in evaluating edge detection~\cite{arbelaez2011contour,dollar2013structured,xie2015holistically,fernandez2018badacost}. 

The input to our evaluation pipeline consists of (1) the probability of each pixel being on edge (fold or occlusion) $p_e$, and (2) a label of each pixel being occlusion or fold.
By thresholding on $p_e$, we first obtain an edge map $E_\tau$ at threshold $\tau$. We denote the occlusion pixels as $O$ and the fold pixels as $F$. We find the intersection $O \cap E_\tau$ and use the same protocol as ~\cite{arbelaez2011contour} to compare it against the ground-truth occlusion $O^*$ and obtain true positive count TF$_o$, false positive count FP$_o$ and false negative count FN$_o$. We follow the same protocol to compare $F \cap E_\tau$ against ground-truth fold $F^*$ and obtain TF$_f$, FP$_f$ and FN$_f$. 

We then calculate the joint counts TF, FP and FN: TP=TF$_o$+TF$_f$, FP=FP$_o$+FP$_f$ and FN=FN$_o$+FN$_f$. 

We iterate through different $\tau$ to obtain the joint counts TF, FP and FN at each threshold to obtain the final ODS/OIS F-score and AP.

\section*{A7: Crossed-out Numbers in Tables}
\label{appendix:crossed_out_numbers}
We made minor quality improvements to the dataset after the camera ready deadline of CVPR 2020, affecting less than 10\% of the images. The crossed-out numbers are those presented in the CVPR camera ready version of this paper~\footnote{\url{https://openaccess.thecvf.com/content_CVPR_2020/papers/Chen_OASIS_A_Large-Scale_Dataset_for_Single_Image_3D_in_the_CVPR_2020_paper.pdf}} and are from the older, obsolete version of the dataset. The publicly released version of OASIS is the new version with the quality improvements.

\end{document}